# Autoencoder-based Semantic Novelty Detection: Towards Dependable AI-based Systems[1]

Andreas Rausch
Institute for Software and Systems Engineering
Technische Universität Clausthal
Clausthal-Zellerfeld, Germany
andreas.rausch@tu-clausthal.de

Azarmidokht Motamedi Sedeh
Institute for Software and Systems Engineering
Technische Universität Clausthal
Clausthal-Zellerfeld, Germany
azarmidokht.motamedi.sedeh@tu-clausthal.de

Meng Zhang
Institute for Software and Systems Engineering
Technische Universität Clausthal
Clausthal-Zellerfeld, Germany
meng.zhang@tu-clausthal.de

*Abstract*— Many autonomous systems, such as driverless taxis, perform safety critical functions. Autonomous systems employ artificial intelligence (AI) techniques, specifically for the environment perception. Engineers cannot completely test or formally verify AI-based autonomous systems. The accuracy of AI-based systems depends on the quality of training data. Thus, novelty detection – identifying data that differ in some respect from the data used for training – becomes a safety measure for system development and operation. In this paper, we propose a new architecture for autoencoder-based semantic novelty detection with two innovations: architectural guidelines for a semantic autoencoder topology and a semantic error calculation as novelty criteria. We demonstrate that such a semantic novelty detection outperforms autoencoder-based novelty detection approaches known from literature by minimizing false negatives.

## I. Introduction

Recently, autonomous systems have achieved success made in many application domains, including autonomous vehicles (AVs), smart home systems, and autonomous financial agents. Autonomous systems are becoming more useful and beneficial for us. As side effect, we – the users – increasingly rely on their services, even in safety-critical applications like driverless taxis [1] [2].

Many recent advancements in the performance of autonomous systems have been made possible by the application of machine learning (ML) techniques [3]. Nowadays, autonomous systems are hybrid *AI-based systems*, integrating classical engineered subsystems combined with subsystems using artificial intelligence (AI) techniques. For instance, the vehicle controllers are classical engineered subsystems whereas the perception subsystems of AVs are nowadays mainly AI-based. Both integrated in an AV perform together safety-critical functions.

For the development of perception subsystems engineers use labeled training data and ML technologies to train an interpretation function. For illustration, a simplified perception task is shown in Fig. 1: The classification of traffic sign images to the corresponding traffic sign classes, i.e., a semantic interpretation of images. Training data map a finite set of input data (images of traffic signs) to correct output information (traffic sign classes). A machine-learned interpretation function is trained using this training data. As ML abstracts from the given training data examples and produces a machine-learned function that can process an infinite number of different images, the resulting machine-learned interpretation function maps any type of image taken by the camera in a real environment to one of the traffic sign classes. During system operation from a safety perspective, an essential question is whether the produced output information $if_{ml}(x)$ of the machine-learned interpretation function $if_{ml}$ processing the current input data $x$ is sufficiently reliable to be safe.

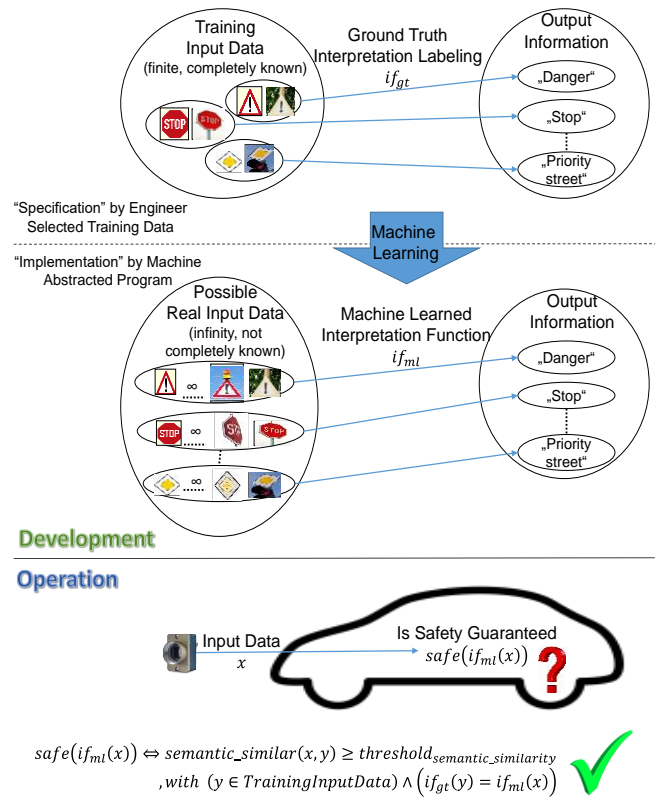

Fig. 1. Operation Time Check of Dependability Requirements

There are differences between the engineering of classical systems and AI-based systems [4]: the classical engineering process for safety critical applications starts with a (semi-) formal requirements specification, which must be complete and correct. While this idealized process is rarely realized to its full extend, the requirements specification is later used as the main input for testing and verification. Instead of requirements specification, a large data collection is used for the development of an AI-based system. The collected data

---

[1] The presented approach has been registered as patent: EP 21 181 014.8 and DE 10 2020 122 735.3, "KI-basierte automatische Erkennung von neuen und relevanten Datensätzen", 31.8.2020.

may have missing information or contain a small percentage of incorrect data samples.

Nevertheless, these AI-based systems have been widely applied to fulfill safety-critical tasks. Product liability regulations impose high standards on manufacturers regarding the safe operation of such systems [5]. Established engineering methods are no longer adequate to guarantee the dependability requirements (safety, security, and privacy) in a cost-efficient way due to many limitations, for instance, they are not able to handle the specific aspects of AI-based systems [1]. Therefore, engineers cannot completely test and verify autonomous systems during development to fully guarantee their dependability requirements in advance. Thus, Aniculaesei et al. [6] have introduced the dependability cage concept. Dependability cages are derived from existing development artifacts and are used to test the system's dependability requirements both during development and during operation.

One essential part of the dependability cage concept (cf. Fig. 3) is the quantitative monitor. In the scenario of Fig. 1, the safety task of the quantitative monitor is intuitively to indicate whether the system is currently processing actual real input data $x$, which are from a reliability perspective similar enough ($semantic\_similar(x, y) \geq threshold_{semantic\_similarity}$) to the (ground truth) training input data ($y$) used for ML so that the produced actual output information of the machine-learned interpretation function $if_{ml}(x)$ can be assumed to be correct and safe. If this is not the case, the AI-based system is possibly in an unsafe state.

Providing a measure for the semantic similarity of input data that serve as an argument for the output information reliability is a challenge ($semantic\_similar(x, y)$). Novelty detection to automatically identify new relevant test data differing from the available training data becomes an interesting approach to realize such a semantic similarity measure. Very promising novelty detection approaches use AI-technologies, such as an autoencoder or generative adversarial network approaches [15].

The basic principle for autoencoder-based novelty detection was introduced by Japkowicz et al. [21]. An autoencoder was trained to minimize the error between an input image and a reconstructed input image. Known images are used to train the autoencoder. After training, the autoencoder was fed with new images. If the difference between the original image and the reconstructed image is higher than a given threshold value, the new image is suspected to be novel. Richter et al. [12] trained an autoencoder with known images of handwritten digits. Their autoencoder-based novelty detection approach could identify images of letters with nonstandard fonts as novel. Amini et al. [14] demonstrated the same with images from known daytime versus novel nighttime driving situations. These "naive" autoencoder-based novelty detection approaches can meaningfully generalize to a family of input images with a common structure and classify input images with different structures as novel data.

However, a significant structural difference in input data does not necessarily correlate with a different output information class. To guarantee the dependability of an AI-based system, we must ensure that there are sufficient comprehensive training input data for each relevant output information, regardless of how large the structural difference in input data is. For instance, as illustrated in Fig. 2, completely different traffic situations (output information) frequently have similar images (input data). On the right-hand side, the AV has to stop and let the pedestrians pass the zebra crossing. On the left-hand side, the AV can pass the zebra crossing.

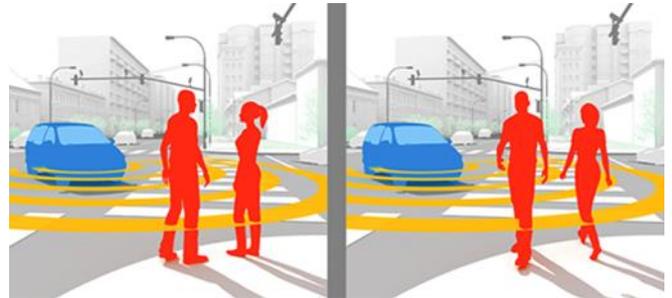

Fig. 2. Similar images of pedestrians represent different situations [31]

AI-based systems are generally known to lack robustness, e.g., in the case that the training data did rarely or completely not cover critical situations. The challenging task is, if a new image represents a data example for an unknown driving situation (relevant new output information), it must be identified as novel, regardless of how similar it is to an existing one.

To mitigate this problem, we propose in this study a semantic novelty detection approach using a new autoencoder architecture that (a) provokes the autoencoder to reconstruct novel input data with a distinct sufficient reconstruction error and (b) calculates the difference between the original and the reconstructed input based on the resulting error of the corresponding output information. We demonstrate that such a sophisticated autoencoder-based semantic novelty detection approach is much more powerful than existing "naive" autoencoder-based novelty detection, such as the approach of Richter et al. [12].

To compare our results with the "naive" autoencoder-based approach, we apply the same setting as Richter et al. using the MNIST data set for our experiments in this paper. As the presented work is part of an ongoing research our results will be later validated in real autonomous driving use cases. More extensive experiments will be published in subsequent publications. However, the results, we present in this paper based on the MNIST-dataset are promising. Once they are validated in real traffic situations, such a semantic novelty can be used to (1) continuously improve the quality of training data for development, (2) to provide a safety argument for the approval of an autonomous system, and (3) to establish an on-board diagnosis system for system operation. In addition, it can be used to identify corner cases and facilitate the de-biasing of training data.

The rest of this paper is structured as follows: Section II introduces a quick overview of the dependability cage monitoring architecture for autonomous systems. Section III provides an overview of related work, particularly AI-based novelty detection. In Section IV, the main research hypotheses are introduced. To evaluate the proposed concepts and research hypotheses, we describe in Section V the evaluation framework, which we have used. In Section VI to VIII, the architecture of the autoencoder-based semantic novelty detection is incrementally developed and evaluated concerning our three research hypotheses. Section IX rounds the paper up with a short conclusion and future work.

## II. DEPENDABILITY CAGE APPROACH—A BIRD'S EYE VIEW

To tackle the challenges of engineering-dependent autonomous systems, the dependability cage concept has been proposed [6] [9] [10] [11] [29]. Dependability cages are derived by engineers from existing development artifacts. The core concept of the approach is a continuous monitoring framework, as shown in Fig. 3.

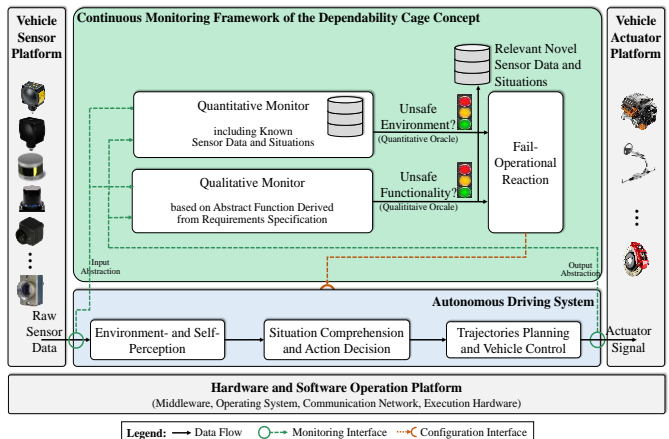

Fig. 3. Continuous monitoring architecture a part of dependability cage

For developing AVs, a common high-level functional architecture has been established, which includes three parts: (1) environment- and self-perception, (2) situation comprehension and action decision, and (3) trajectories planning and vehicle control [7] [8] [13].

The monitor framework focuses on two issues: (a) shows the system correct behavior in terms of dependability requirements (*Qualitative Monitor*), and (b) operates the system in a situation and environment that has been trained or tested during development (*Quantitative Monitor*) (cf. Fig. 3).

Both monitors require consistent and abstract data access to the system under consideration, with the help of *input and output abstraction*. They depict the interface between the autonomous system and the two monitors. Both *input* and *output abstraction* components use defined interfaces to access the autonomous system's data and transform it into abstract representations. Abstract representation types and values are derived from requirements specification and dependability criteria of the autonomous system.

The qualitative monitor evaluates the correctness and safety of system behavior under the assumption that the system operates in a situation and environment that conform to the requirements specification, and it consists of an abstract behavior function and a conformity oracle. The abstract behavior function calculates a set of correct and safe actions in real time for the system in the current abstract situation. The conformity oracle compares the output abstraction with the set of correct and safe abstract actions from the abstract behavior function.

The quantitative monitor observes the encountered abstract situations. For each situation, the monitor evaluates in real time if the encountered abstract situation is already known from development. A knowledge base provides information about tested situations on an abstract level. A canonical representation of abstract situations is used in this study. These canonic abstract situations are considered as unique situations.

If one of the monitors detects incorrect and unsafe system behavior, safety measures must be initiated to guarantee dependability requirements. The *Fail-Operational Reaction* component has to transfer the corrupted system into a safe state with acceptable risk. For instance, one fail-operational reaction can be a graceful degradation of the autonomous system, as described in [25].

Moreover, in the case that safety measures are initiated, system data will be automatically recorded. The recorded data can be transferred back into system development to identify and to resolve previously unknown faulty system behavior as well as to extend the scope of system tests. The qualitative monitor has already been successfully applied and evaluated using an industrial prototype of a highway pilot and data collected from road tests on German highways [9] [11].

To realize a quantitative monitor, we need an efficient and effective solution to identify unknown sensor input data. As already mentioned, we claim and will show in the following sections that our proposed semantic novelty detection approach can be used to successfully realize a quantitative monitor for a perception system.

## III. STATE OF THE ART IN NOVELTY DETECTION

Novelty detection as a classification task to determine whether newly observed data differ from the available training and test data has been widely used in different fields. Different approaches of novelty detection have been investigated [15] [16] [18]. Novelty detection is closely related to anomaly detection, which typically refers to the detection of samples during inference that do not conform to an expected normal behavior [17]. If enough diverse training data is available any anomaly is a novelty, but not any novelty is necessarily an anomaly. Hence, novelty detection is the more general and such safer approach for us. Note, in automotive novelty detection is also known as corner case detection.

One well-known classification of novelty detection approaches is proposed in [15], consisting of five taxonomies: probabilistic, distance-based, domain-based, reconstruction-based, and information theoretic. In our work, we focus on the taxonomy of reconstruction-based approaches, which are often used on safety-critical applications and cover approaches based on diverse neural networks (NNs) such as MLP, SVM, and auto-associator, as well as oscillatory and habituation-based networks [19].

The auto-associator approaches, particularly referring to autoencoder approaches, have gained our attention in the research because of its strong ability implicitly to learn the underlying characteristics of an input dataset without any prior knowledge or assumptions [19]. The autoencoder approaches automatically learn data abstraction on different implicit representation levels by attempting to regenerate the system output as the same as the originally given input, such as camera images, which have been investigated in different studies [22] [23] [24].

For example, Byungho et al. identified three critical properties of autoencoder-based novelty detection: (a) production of the same output vector with infinite input vectors, (b) output-constrained hyperplane when using bounded activation functions, and (c) hyperplane located in the vicinity of training pattern by the minimizing error function [20]. Diaz et al. found that kernel regression mapping is better suited than the least squares mapping for the autoencoder networks of novelty detection [22]. Japlowicz et al. [21] focused on the idea of setting the threshold as some function of the largest reconstruction error for any single training data point. Based on this, Richter et al. [12] selected a high percentile of the cumulative distribution of losses on training datasets as their threshold to make it more robust in detecting novel instances with different input data structures. Their experimental setting – handwritten digits versus (novel) letter images – serves as starting point for our work. To improve the criterion of novelty detection, Manevitz et al. investigated autoencoders on different levels of error and determined an optimal real multiple of the standard deviation of the average reconstruction error as a better threshold. Their evaluation was based on an application for document classification [26].

As already introduced in the previous sections, the aim of our work is to realize a quantitative monitor as part of the dependability cage by using a new improved autoencoder-based approach. The autoencoder-based approach by Richter et al. [12] is used as a reference in our study, which details are introduced in Section V–VIII. By introducing a new semantic novelty detection approach using a new autoencoder architecture, we will improve the results of Richter et al.

## IV. RESEARCH HYPOTHESIS OF THIS WORK

To realize the previously mentioned task of quantitative monitoring, we propose an autoencoder-based approach of semantic novelty detection during real-time system operation to classify a current situation into (a) tested and known or (b) novel and unknown. In the case of (b), the corresponding raw data will be recorded for further development of the monitored autonomous system. Moreover, the classification may also lead to a (a) safe or (b) unsafe state decision in the system and a corresponding fail-operational reaction. It can be concluded that from a viewpoint of safety it is important to detect all novel situations. Otherwise, the system decides to operate in an unlearned environment, which could lead to a safety-critical situation.

In Section III, an autoencoder-based novelty detection approach by Richter et al. [12] was briefly introduced, in which compared to the images of the MNIST handwritten digits dataset [27], the images of letters with nonstandard fonts from the notMNIST dataset [28] were successfully classified as novel. To compare our results with their approach, we reproduced the results of Richter et al. in Section V. We focused particularly on their example of detecting letter images as novel. Our reproduced results showed that Richter's approach works well in identifying novel data that have different input data structures compared to the MNIST-dataset, which is used to train the autoencoder.

In this paper, we argue that their success depends on major structural differences in the input data: images of digits and letters. Therefore, we modified experiment of Richter et al. such that we randomly excluded some classes of digits from the training set of the autoencoder to figure out whether the autoencoder can still detect these unknown digits as novel. Therefore, in Section V, we will present a slightly modified dataset, containing seven classes of digits for training, and the other three classes of digits as well as nine classes of letters remain as novel data.

We expect that Richter's approach will fail once the novel data have a minor structural, but a major semantic difference. Consequently, images will be detected as known instead of novel, producing a high rate of false negative errors. In the subsequent section, Richter's approach is taken as a reference and named as the "naive" autoencoder-based novelty detection, which failure regarding the false negative errors will be furtherly proved in Section VI. Additionally, a normal classification neural network for novelty detection will be introduced in Section V, which is named as simple classification approach in this study and serves as a lower bound benchmark solution in our evaluation.

To minimize the false negative errors, we improve autoencoder-based novelty detection in two steps. The first one is based on a semantic topology, which is called semi-semantic novelty detection. In this approach, the semantics of the monitored system's output, which is a classification system, is considered. We force the autoencoder to reflect the semantics in its topology by applying three rules (cf. Section VII). In a second step, to realize a so-called fully-semantic novelty detection, we reuse the topology of semi-semantic novelty detection, but the novelty detection criterion is replaced by a semantic loss instead of the original mean squared error (MSE). With this, our research hypotheses for upcoming sections are as follows:

**HYPO 1**: "Naive" autoencoder-based novelty detection does not work well for similar structural input data with a different semantic interpretation.

**HYPO 2:** Autoencoder-based semi-semantic novelty detection works better than the "naive" approach.

**HYPO 3:** Autoencoder-based fully-semantic novelty detection can find novel data completely.

The first hypothesis is evaluated in Section VI by using the simple classification network mentioned above as a lower bound benchmark solution. The last two hypotheses are evaluated, respectively, within benchmarks with the "naive" approach to evaluate the performance of both improved approaches for minimizing false negative errors in the proposed novelty detection approach (cf. Section VII and Section VIII). Our results show that both approaches have significantly fewer false negatives than the "naive" approach, and the fully-semantic approach can find all novel data in our experimental setup.

The following section introduces our concrete evaluation framework, including the repetition of the original experiment of Richter et al. [12], a slight change to the applied dataset of Richter et al. with respect to our HYPO1, and the establishment of the mentioned lower bound benchmark solution based on the normal classification neural network with the usage of the changed dataset.

## V. EVALUATION FRAMEWORK FOR THIS WORK

To evaluate our proposed semantic novelty detection approaches, we developed an evaluation framework based on Keras in the TensorFlow framework. The mentioned novelty detection approaches ("naive", semi-semantic, and fully-semantic) were implemented as ROS2 [30] nodes and serve as part of the quantitative monitor (cf. Fig. 3). For a comparable and comprehensive evaluation, we firstly reproduced the results of Richter et al. [12] by training a feedforward autoencoder with configurations in TABLE I.

The autoencoder aims to minimize the mean squared pixel-wise deviations between raw and reconstructed images during training. The $99^{th}$ percentile of the empirical cumulative distribution of reconstruction errors for the training set (cf. TABLE I) was decided as the threshold to identify known and unknown data. In the case of an image, if the reconstruction error falls below this threshold, the input image is classified as known, otherwise as unknown (novel).

TABLE I. TRAINING SETUP OF AUTOENCODER IN "NAIVE" NOVELTY DETECTION

| Setup of training Autoencoder in "Naive" Novelty Detection Approach | | |
|---|---|---|
| Autoencoder Topology | 784-50-50-50-784 | |
| Hyperparameters | Batch Size | 1024 |
| | Epochs | 3000 |
| | Learning Rate | 0.01 |
| | Activation Function | sigmoid |
| Training set | Randomly selected 30.000 digit images from MNIST-dataset | |
| Test set | Randomly selected 12.500 digit images from MNIST-dataset | |

One hundred randomly selected digit images were taken as the dataset of known classes (Known digits in Fig. 4), and additional one hundred letter images from the notMNIST dataset (Letters in Fig. 4) were taken as novel data. Our result of novelty detection (cf. Fig. 4) is similar to the work of Richter et al. [12] and shows that the autoencoder detects all digit images as known (black graph line) and all letter images as unknown (red graph line). Thus, we could reproduce the results of [12].

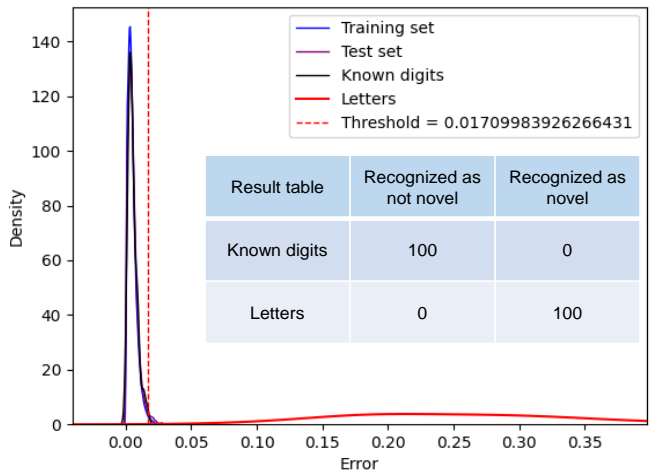

Fig. 4. Reproduced results of "naive" novelty detection approach of [12]

To prove our HYPO 1-3 above regarding the work of Richter et al in [12], namely that the "naive" autoencoder approach fails once the novel input data have a minor structural but a major semantic difference and that our new semantic autoencoder approach is even still convincing in those cases, we modified the experimental setting of the dataset. We randomly excluded three digit classes ("2," "3," and "5") from the training and test set. The other seven classes of digits remained in the known training data. Thus, novel input data consists now out of three classes of digits and nine classes of letters. The following experiments have all used the same dataset shown in TABLE II.

Based on the resulting probability, a simple novelty detection can be performed. We used the $99^{th}$ percentile as the threshold, which means that if the probability for one of the classes is higher than 0.99, the input image is categorized into this class. Otherwise, if none of the probabilities is higher than the threshold, the input image cannot be classified into any known class and thus is identified as novel.

The simple classification network already mentioned in Section IV was built sequentially with three convolutional layers as follows: one ReLU convolutional layers with 28 filters and kernel size of (3, 3) with the same padding, and two ReLU convolutional layers with 64 filters and kernel size of (3, 3) with the same padding. Each of them was followed by max pooling with a pool size of (2, 2), dropout of 0.2, and batch normalization. Afterward, this convolutional network was followed by two fully connected ReLU layers with each including 256 and 128 nodes, as well as a ReLU output layer with 7 nodes according to the number of known classes. We used the Adam optimizer and attempted to minimize the categorical cross-entropy loss during training. To maintain evaluation authenticity, we used the same number of images for training and testing results (cf. TABLE II).

TABLE II. USED DATASET FOR THE EVALUATION BENCHMARKS

| Training and Test of Autoencoder | |
|---|---|
| Training set | 30.000 randomly selected images of known digits ($KN \in \{0, 1, 4, 6, 7, 8, 9\}$) |
| Test set | 12.500 randomly selected images of known digits ($KN \in \{0, 1, 4, 6, 7, 8, 9\}$) |
| Performance Evaluation of Novelty Detection | |
| Known digits | 100 x randomly selected images of known digits ($KN \in \{0, 1, 4, 6, 7, 8, 9\}$) |
| Unknown digits | 100 x randomly selected images of unknown digits ($UN \in \{2, 3, 5\}$) |
| Letters | 100 x randomly selected images of (unknown) letters ($L \in \{A, B, C, D, F, G, H, I, J\}$) |

The performance of the novelty detection based on the simple classification network is shown in Fig. 5. As introduced before, it will serve as a lower bound benchmark for the following experiments. It is noted that for this simple classification approach, we could not show the density graph of the reconstruction error because the classification approach is not based on the principle of image reconstruction. As shown in Fig. 5, almost all known digits were classified as not novel. Most of the unknown digits were classified as novel, but most of the letters were identified as not novel.

| Result table | Recognized as not novel | Recognized as novel |
|---|---|---|
| Known digits | 97 | 3 |
| Unknown digits | 14 | 86 |
| Letters | 69 | 31 |

Fig. 5. Performance of simple classification network for novelty detection

In the following three sections, we will use this evaluation framework and the lower bound benchmark solution based on the simple classification network to prove HYPO 1-3 and show the performance of the autoencoder approaches within following experiments: (1) "naive" novelty detection vs. classification network (cf. Section VI), (2) semi-semantic novelty detection vs. "naive" novelty detection (cf. Section VII), and (3) semi-semantic novelty detection vs. fully-semantic novelty detection (cf. Section VIII).

## VI. "NAIVE" AUTOENCODER-BASED NOVELTY DETECTION

### A. Architecture

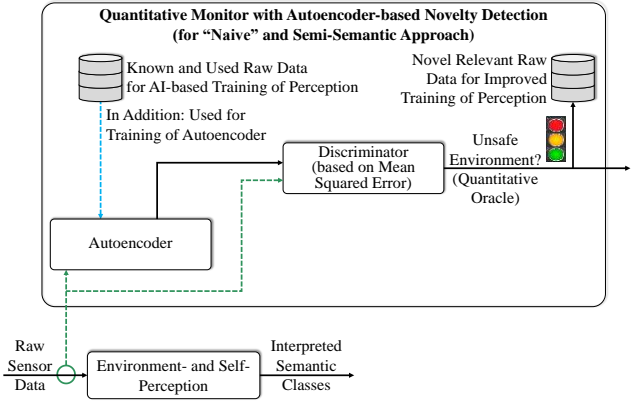

Fig. 6. Quantitative monitor with autoencoder-based novelty detection

To evaluate the performance of the "naive" autoencoder-based novelty detection approach, we used a quantitative monitor (cf. Fig. 3), whereas the input abstraction of the quantitative monitor is realized by the autoencoder component. The quantitative monitor itself in Fig. 3 is represented by a discriminator component, as shown in Fig. 6.

The autoencoder component here refers to the "naive" one (following the Richter's "naive" approach). It is trained with the same data used for the training of the normal digit classification network, which serves as a toy sample for the perception part of an autonomous system. Thus, the autoencoder has abstract knowledge of known data from training and testing.

The configuration of the "naive" autoencoder is described in TABLE I, which is generated from our reproduced results of [12]. The discriminator component compares the output of the autoencoder, which is the reconstructed input by the autoencoder, with the inputs of raw sensor data to compute a reconstruction error in terms of MSE (mean squared error).

The autoencoder is trained to successfully reconstruct known (from training and test) input data, which was in [12] assumed to have a significantly lower MSE than novel input data. Based on the computed threshold mentioned in Section V ($99^{th}$ percentile of the empirical cumulative distribution of reconstruction errors), the discriminator determines whether the input sensor data are known or novel. In the case of novelty, novel relevant data will be recorded to (re-)train and thus improve the perception part of an autonomous system. After training the perception part, the new relevant data will be taken as known raw data to (re-)train the autoencoder.

### B. Evaluation

Since Richter et al. [12] did not consider the unknown digits (cf. Fig. 4), the performance of their approach cannot be compared with that of the simple classification approach (cf. Fig. 5). Therefore, we had to use our refined evaluation framework with Richter's "naive" autoencoder-based novelty detection.

The results of our evaluation shown in Fig. 7 confirmed HYPO 1. Using our evaluation framework with the "naive" autoencoder (cf. TABLE II) illustrates that the "naive" autoencoder-based approach cannot identify the structural similarity but semantic difference between the unknown digits as novel (green graph line in Fig. 7). The results in Fig. 7 show that 83% of the unknown digits are not identified as novel. The remaining results were comparable to the original evaluation by Richter et al. and were also comparable to our reconstruction of the evaluation, as shown in Fig. 4.

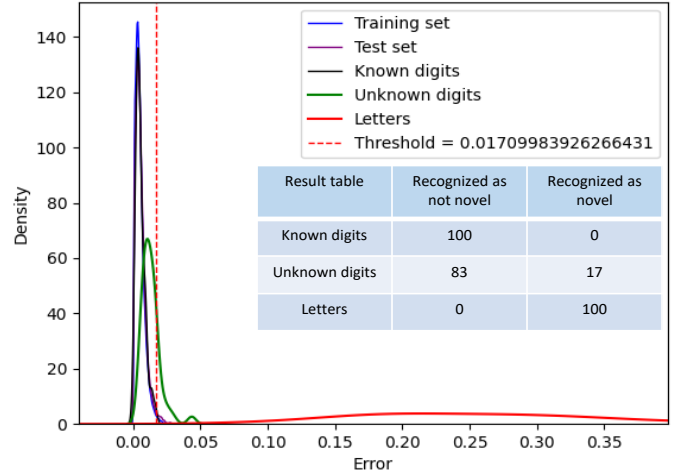

Fig. 7. Performance of "naive" autoencoder-based novelty detection

Overall, the "naive" autoencoder-based novelty detection only performs well when there is a strong structural difference in input images (Known digits vs. Letters). When there are minor structural differences in input images (Known digits vs. Unknown digits), the "naive" autoencoder-based novelty detection showed a poor performance.

In contrast, the performance of the simple classification approach for novelty detection performed well when there is a minor structural difference (Unknown digits) and poorly in strong structural difference (Letters) (cf. Fig. 5). Hence, a promising novelty detection approach is required to discover novel data when there are structural and semantic differences in input data. This helps the quantitative monitor reduce the number of false negatives (unknown classes of digits resp. letters that are not recognized as novel).

## VII. AUTOENCODER-BASED SEMI-SEMANTIC NOVELTY DETECTION

### A. Architecture

To minimize these false negative errors, we proposed our autoencoder-based semi-semantic approach and used it in the quantitative monitor. To enable a reliable novelty detection under the consideration of correlation between the degree of structural difference in the input data and different semantic interpretations, the topology of the autoencoder is derived from the parameters of the semantics of the monitored system's output in this approach. Since only the topology of autoencoder (component *Autoencoder*) was changed to a so-called *Semantic Topology*, the architecture stays the same (cf. Fig. 6).

In our work, the semantics of the monitored system's output is a classification system comprise different classes. To ensure the autoencoder reflects this semantics in its topology, we postulate the following rule for the number of neurons in the middle layer (the latent space, cf. Fig. 8):

*1st Rule for a semantic topology-based autoencoder:*
$[\log_2 n] \leq f \leq n$;

$f$: Number of neurons in the middle layer.

$n$: Number of known classes.

By applying this rule, we ensured the autoencoder finds an abstraction of classes to be identified in its topology. Therefore, when an input image does not belong to one of the output classes (thus a novel image) the autoencoder reconstructs this input image with a distinct reconstruction error.

In addition to the number of neurons in the middle layer, we postulate two other rules. Note that we also consider input and output for these rules. Therefore, we add "2" at each side of inequality of the next rule:

*2nd Rule for a semantic topology-based autoencoder:*

$$(2 * \lceil \log_2 n \rceil) - 1 + 2 \leq l \leq (2 * \lceil \log_2 n \rceil) + 1 + 2,$$

$l$: Number of layers, including input and output layer.

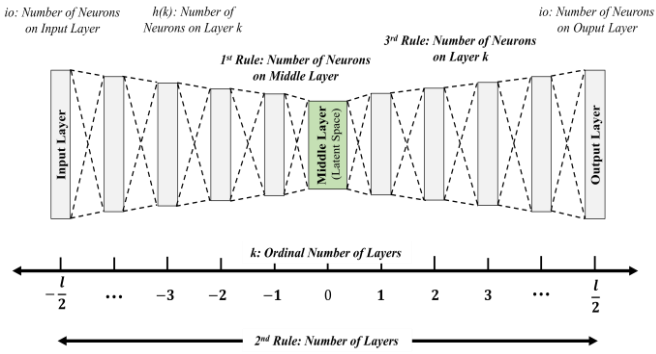

Fig. 8. Rules for semantic topology of the autoencoder

For the last rule, we need an ordinal number for each layer. We defined the middle layer (latent space) to be the layer with the ordinal number 0. The ordinal number $k$ of a layer is the distance to the middle layer. This means that the next layer on the left-hand side to the middle layer has the ordinal number $-1$, the next one on the right-hand side has the ordinal number 1, and so on. We considered the input and output layer to be the layers with the lowest (input layer) and the largest (output layer) ordinal numbers.

*3rd Rule for a semantic topology-based autoencoder:*

$h(k) \coloneqq$
$$\begin{cases} io, \text{with } k = \pm \lfloor l/2 \rfloor \text{ (input or output layer)} \\ h(k) \text{ derived from 1st rule, with } k = 0 \text{ (middle layer)} \\ 2^{\lceil \log_2 io \rceil - \sum_{i=0}^{\lfloor l/2 \rfloor - |k| - 1} i} \geq h(k) \geq 2^{\lceil \log_2 io \rceil - \sum_{i=0}^{\lfloor l/2 \rfloor - |k| - 1} 2^i} \\ \quad , otherwise \end{cases}$$

$k$: Ordinal number of a layer.

$h(k)$: Resulting number of neurons in layer $k$.

$io$: Number of neurons in the input or output layer.

By applying these two additional rules, we ensured that the autoencoder shows an exponential packaging of the relevant features for the semantics of the monitored system's output. We ensured that the autoencoder quickly focuses on the relevant semantic features. The proposed quantitative monitor with the autoencoder-based semi-semantic novelty detection is then compared to the "naive" autoencoder-based approach, which is discussed in greater detail in the following section.

## B. Evaluation

As previously mentioned, three digits in the MNIST-dataset were omitted in the evaluation framework. Thus, the number of classes to be identified had been reduced to seven. Following the 1st rule, the number of neurons in the latent space should stay between three and seven. As a first attempt, we only used seven neurons in the latent space in the original autoencoder. The resulting topologies of the autoencoder were 784-50-7-50-784 neurons, hence the three hidden sigmoid layers had 50-7-50 neurons and the input and output layer had 784 neurons, due to the image size. The rest of the evaluation setting remained constant.

Fig. 9 demonstrates the evaluation result of the first rule of semi-semantic autoencoder, which has a better performance than the "naive" autoencoder (cf. Fig. 7) but does not perform better than the simple classification approach (cf. Fig. 5) regarding the identification of unknown digits.

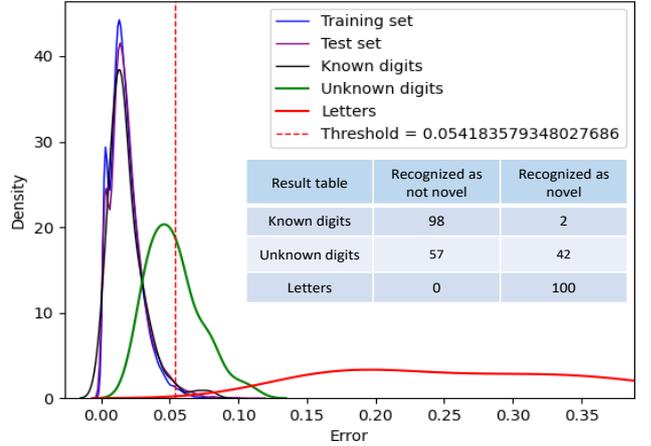

Fig. 9. First improvement of novelty detection with a semantic topology

The 2nd and 3rd rules provide a design space of a semi-semantic topology-based autoencoder for novelty detection. Experimentally, we discovered the best topology for the given novelty detection task within the design space: an autoencoder with five hidden layers, respectively, with the neuron numbers of 512, 128, 7, 128, 512 for these five layers, which evaluation result is better than the "naive" autoencoder-based novelty detection as well as the simple classification network, as shown in Fig. 10. Overall, by using our three rules, we derived an autoencoder-based semi-semantic novelty detection, which performs better than the "naive" autoencoder approach and thus confirms the HYPO 2 (cf. Section IV).

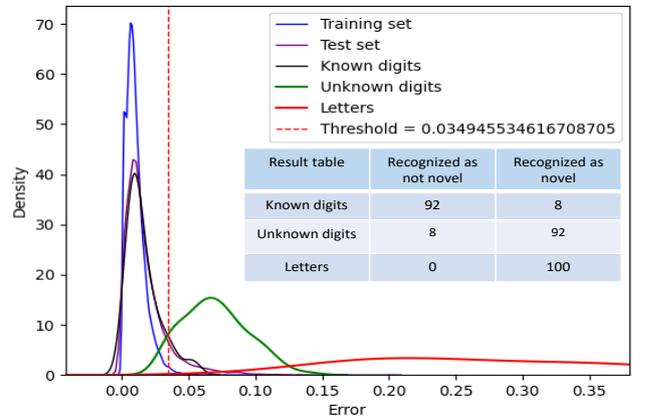

Fig. 10. Applying the rules: novelty detection with a semantic topology

## VIII. AUTOENCODER-BASED FULLY-SEMANTIC NOVELTY DETECTION

### A. Architecture

Based on the evaluation results in Section VII, the semi-semantic approach with semantic topology of the autoencoder has shown a great performance for minimizing the false negative errors in novelty detection. During the evaluation, we found that the current classification criterion of the threshold based on MSE could not segment known and novel input data, which are similarly structured but have different semantic meanings, e.g., in the case of known digits versus unknown digits (cf. Fig. 10). This could also be another reason for the false negative errors in addition to the topology of the autoencoder.

Since the current classification criterion for novelty detection is the threshold solely based on the reconstruction MSE of raw input data, it means that in the case of camera image data, the threshold is determined based on pixel-wise deviations between the raw and reconstructed image without any semantic meaning. In this case, the false negative errors due to similarly structured but semantically different novel data are obviously remaining.

With this hindsight, we propose our next improvement: autoencoder-based fully-semantic novelty detection based on a semantic error instead of the threshold based on MSE. The architecture of the quantitative monitor using this approach is shown in Fig. 11.

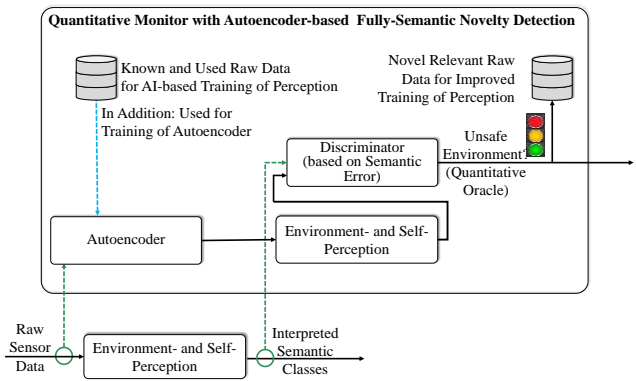

Fig. 11. Quantitative monitor with fully-semantic novelty detection

Compared to the semi-semantic novelty detection, the component *Autoencoder* with the semantic topology (cf. Section VII) remains the same. The output of the autoencoder (reconstructed raw sensor data) is forwarded to the environment- and self-perception component with the identical interpretation mechanism acting as the perception part of the monitored autonomous system, which delivers its own interpreted semantic classes as output (cf. Fig. 11), such as the object list in the case of the autonomous driving system on the vehicle. These interpreted semantic classes are used as inputs of the component *Discriminator* in the quantitative monitor. The discriminator detects novel input by comparing both identified semantic classes, delivered by the components environment- and self-perception, respectively, in the autonomous system and the quantitative monitor.

Thus, it can be understood that the quantitative monitor with the fully-semantic approach has novelty detection on two different levels. First, the identical component *Environment- and Self-perception* classifies the reconstructed raw data provided by the autoencoder. Second, both sets of interpreted semantic classes are compared within the component *Discriminator* on a higher semantic level due to the consideration of integrated semantic errors. Thus, the false negative errors due to similarly structured but semantically different novel data can be minimized.

### B. Evaluation

Based on the architecture, a semi-semantic autoencoder and a classification network, deployed in the component *Environment- and Self-perception,* are required in our evaluation experiment. To find the best result, we used the classification network built for the benchmark mentioned in Section V as a sample component for environment- and self-perception. Additionally, the semi-semantic autoencoder with five hidden layers (512-128-7-128-512) described in Section VII is also used here together with the classification network to build an experimental environment for the evaluation of the fully-semantic approach for novelty detection.

The classification network classifies the input data based on a minimal permitted probability of 99%, which is used in this study as the threshold for the classification task (cf. Section V). The autoencoder reconstructs the raw input data and sends the reconstructed data to the classification network. With the same procedure, the classification network calculates the class of the autoencoder's output (reconstructed input) as well as the class of the original raw input data. If the identified classes of both candidates, including the raw input data and the reconstructed input data, are the same, the input data are identified as known. If there is a difference between these two candidates, the input is novel.

As the results of this approach depend on the classification process, we could not build a density diagram. However, the results shown in Fig. 12 illustrate a perfect novelty detection, which satisfies our expectation of minimizing false negatives and thus confirms the HYPO 3 (cf. Section IV). However, the false positives (known image classes recognized as novel) increase from 8% (cf. Fig. 10) to 20% (cf. Fig. 12). Nevertheless, from a safety perspective, false positive errors are less critical than false negative errors in novelty detection.

| Result table | Recognized as not novel | Recognized as novel |
|---|---|---|
| Known digits | 80 | 20 |
| Unknown digits | 0 | 100 |
| Letters | 0 | 100 |

Fig. 12. Performance of applying the fully-semantic novelty detection

## IX. SUMMARY AND OUTLOOK

A great progress in AI-based systems, such as on the application of AVs, has been achieved due to an increased availability of vast amounts of training data for the underlying ML approaches. ML is generally known to lack robustness, e.g., in the case that the training data did rarely or completely not cover critical situations. Nevertheless, many of these AI-based systems have been applied to fulfill safety-critical tasks such as in the case of AVs. Engineers cannot completely test and verify autonomous systems in advance during development due to many reasons. For instance, they are not able to foresee all situations, which an AI-based system will have to manage. Thus, novelty detection for automatically identifying new relevant data differing from the available training data becomes an interesting approach to improve the robustness of AI-based systems and thereby improve their safety.

In this paper, we successfully reproduced the results of a "naive" autoencoder-based novelty detection approach published in [12]. Based on a benchmark with a simple classification neural network, we demonstrated that such a "naive" approach is incapable of minimizing false negatives, when novel and known input data are similarly structured but semantically different.

To reduce the number of false negatives, we proposed a semi-semantic approach, which uses a semantic topology to ensure that the autoencoder reflects the semantics of the monitored system's output using several derived rules. These architectural rules have been validated with test samples under our observation. In our experiments, the semi-semantic approach reduced the percentage of false negatives in unknown digits from 83% to 8%. However, the three rules have not been extensively evaluated, validated and adjusted. Further ongoing research is therefore required.

Moreover, we discovered that the classification criterion for novelty detection, which is a threshold solely based on MSE, limits the semi-semantic approach's ability to reduce the number of false negatives. To overcome this limitation, we proposed a fully-semantic approach, which relies on a semantic loss instead of on the original MSE. The semantic loss is realized by a new architectural structure: Instead of comparing with the input data, the loss is calculated based on the output of the AI-based system, in our case referring to the digit classification function, feed with the original input data as well as with the reconstructed input data. Thus, the fully-semantic approach does not only detect novelty on the input data level but also on a higher semantic level. Consequently, the critical cases of novelty detection regarding similarly structured but semantically different novel data were eliminated. Based on the comparison of the fully semantic and the semi-semantic approaches in our experiments, the percentage of false negative errors for the classification of unknown digits was reduced from 8% to 0%.

The results of our paper demonstrated that semantic novelty detection is able to prevent false negative errors, especially for structurally similar but semantically different input data. However, false positive errors increase. From a safety perspective, false positive errors are less critical than false negative errors in novelty detection.

In this paper, datasets such as handwritten digit images and letters are used in experiments, which are different from data used in the real-world domain for AVs. Another limitation is that our test samples and experiments do not allow generalizing the presented results (architectural rules and structure). In the future, real traffic images are required to be used to evaluate the proposed semantic novelty detection. The postulated architectural rules have to be validated with more experiments. Moreover, other network types should be evaluated. This will lead us to a semantic novelty detection approach for the quantitative monitor to efficiently improve the quality of training data, and, thus, the safety of the monitored AI-based systems, in the long run, e.g., in the case of AVs. Thereby, it could serve as a safety argument for the approval of AI-based systems.

VALIDATION AND ACCKNOWLEDGEMENTS

We thank Yunsu Cho, who came to our research team as a new employed research assistant and Ph.D. candidate. Based on a pre final version of this paper, she was able to reproduce our whole experiments and thereby had independently validated our research results. Moreover, we thank our internal reviewer Rüdiger Ehlers and Adina Aniculaesei. They provided many helpful comments to improve this paper. And last but not least, we thank all anonymous referees for their useful suggestions.